\documentclass[a4paper]{article}

\usepackage{INTERSPEECH2020}

\title{User-Initiated Repetition-Based Recovery in Multi-Utterance Dialogue Systems}
\name{Hoang Long Nguyen, Vincent Renkens, Joris Pelemans, Srividya Pranavi Potharaju, \linebreak Anil Kumar Nalamalapu, Murat Akbacak}
\address{Apple, 1 Apple Park Way \\
  Cupertino, California}
\email{\{romanhoangnguyen\_long, vrenkens, jpelemans, spotharaju, anilkumar\_nalamalapu,  makbacak\}@apple.com}

\begin{document}

\maketitle
\begin{abstract}

Recognition errors are common in human communication. Similar errors often lead to unwanted behaviour in dialogue systems or virtual assistants. In human communication, we can recover from them by repeating misrecognized words or phrases; however in human-machine communication this recovery mechanism is not available. In this paper, we attempt to bridge this gap and present a system that allows a user to correct speech recognition errors in a virtual assistant by repeating misunderstood words. When a user repeats part of the phrase the system rewrites the original query to incorporate the correction. This rewrite allows the virtual assistant to understand the original query successfully. We present an end-to-end 2-step attention pointer network that can generate the the rewritten query by merging together the incorrectly understood utterance with the correction follow-up. We evaluate the model on data collected for this task and compare the proposed model to a rule-based baseline and a standard pointer network. We show that rewriting the original query is an effective way to handle repetition-based recovery and that the proposed model outperforms the rule based baseline, reducing Word Error Rate by 19\% relative at 2\% False Alarm Rate on annotated data. 
\end{abstract}

\section{Introduction}

Virtual assistants that use Automatic Speech Recognition (ASR) rely on the accuracy of the transcript to understand the user's intent. However, even though ASR accuracy has improved, recognition errors are common. Errors typically occur due to both noise and the ambiguity of language. Xiong et al. \cite{xiong2016achieving} found the error rate of professional transcribers to be 5.9\% on Switchboard, which is a dataset of conversational speech. Specifically, for out-of-vocabulary words, like names, ASR systems often make mistakes. Recognition errors lead to misunderstandings of the user's intent, which results in unwanted behaviour. Recognition errors can be mitigated by detecting that an error has occurred \cite{ogawa2017error, fayolle2010crf}. If this is the case, the virtual assistant might initiate a recovery strategy. Typical recovery strategies include asking clarifying questions or asking the user to repeat a certain part of the request \cite{skantze2008galatea}.\\

If the system fails to detect a recognition error and asks the user to correct, it may respond nonsensically or take an action the user did not intend. In response, the user might try to alleviate the misunderstanding. The most common way to do this is repeating (part of) the utterance \cite{swerts2000corrections}. We call this repetition-based recovery. Substantial work exists on detecting if such a repetition-based correction occurred using prosody \cite{litman2006characterizing, hirschberg2004prosodic, stifelman1993user}, n-best hypothesis overlap \cite{kitaoka2003detection}, phonetic distances \cite{lopes2015detecting} or a combination of the above \cite{skantze2004early}. \\

In this paper, we focus on automatically correcting the error that was made through Query Rewrite (QR). Query Rewrite is used in information retrieval systems \cite{musa2019answering, DBLP:journals/corr/abs-1809-02922} and smart assistants \cite{ChenQR, RoshanGhias2020PersonalizedQR, Rastogi2019ScalingMD}. We generate a corrected transcription of the incorrectly understood utterance. A Natural Language Understanding (NLU) system can then parse this rewritten query so that the system can recover from the error. QR is a generalizable way of handling repetition-based recovery, because it makes no assumptions about a specific grammar the user should be following. This is especially important for virtual assistants that handle a wide range of requests. Previously proposed systems for repetition-based recovery \cite{sagawa2004correction, lopes2015detecting} assume a specific grammar or task making them hard to use for a virtual assistant. \\

We propose a model that can generate a rewritten query by merging together the incorrect first utterance and the correction follow-up utterance. We can then use the rewritten query to re-estimate the user's intent and act upon it.
The inputs to the model are Acoustic Neighbour Embeddings (ANE) \cite{jeon2020acoustic} of the words in the utterances. ANEs are embeddings that are trained in such a way that similar sounding words are close to each other in the embeddings space. The ANE model takes in a word as a sequence of graphemes and produces a word embedding. With these embeddings, the model can infer which words sound similar and thus could have been confused by the ASR. 
Wang et al. \cite{DBLP:conf/interspeech/WangDLLAL20} rewrite single turn utterances by taking phonetic inputs from the ASR model to handle entity recognition errors.

Our proposed model is an encoder-decoder \cite{sutskever2014sequence} network. The decoder is a modified pointer network \cite{vinyals2015pointer} that takes two input sequences, one for the incorrect first utterance and one for the correction follow-up utterance. A pointer network is an attention based \cite{bahdanau2014neural} model that outputs words from a variable-sized dictionary. A pointer network is necessary because the model only considers words occurring in the two input utterances as potential outputs. We modify the standard pointer network to a 2-Step attention pointer network, which is tailored to this problem. The pointer network first attends over the first turn, then attends over the second turn and finally selects between the two attention results.\\

The model is trained using synthetic data generated based on a transcribed speech dataset. Using the ASR transcription and the reference transcription, a repetition-based recovery is generated. This data generation method allows the rewrite model to be trained using existing commonly available data resources. 

\noindent The contributions of this paper are threefold:
\begin{enumerate}
\item We propose QR as a generalizable way of handling repetition-based recovery.
\item We propose 2-Step Attention (SA2) pointer network for this task.
\item We propose a method to generate training data based on transcribed speech. 
\end{enumerate}

The remainder of this paper is organised as follows. First, in Section \ref{sec:model}, we describe the proposed model, and in section \ref{sec:training}, we explain how we generate training data to train this model. In Section \ref{sec:baseline}, we describe the models we used as baselines. Next, in Section \ref{sec:experiments}, we present the data we used for the experiments, the metrics we used to measure the performance and the model hyper-parameters. In Section \ref{sec:results}, we discuss the results. Finally, in Sections \ref{sec:future} and \ref{sec:conclusion}, we formulate future work and conclusions. 

\section{Model}
\label{sec:model}

\begin{figure}
\includegraphics[width=\linewidth]{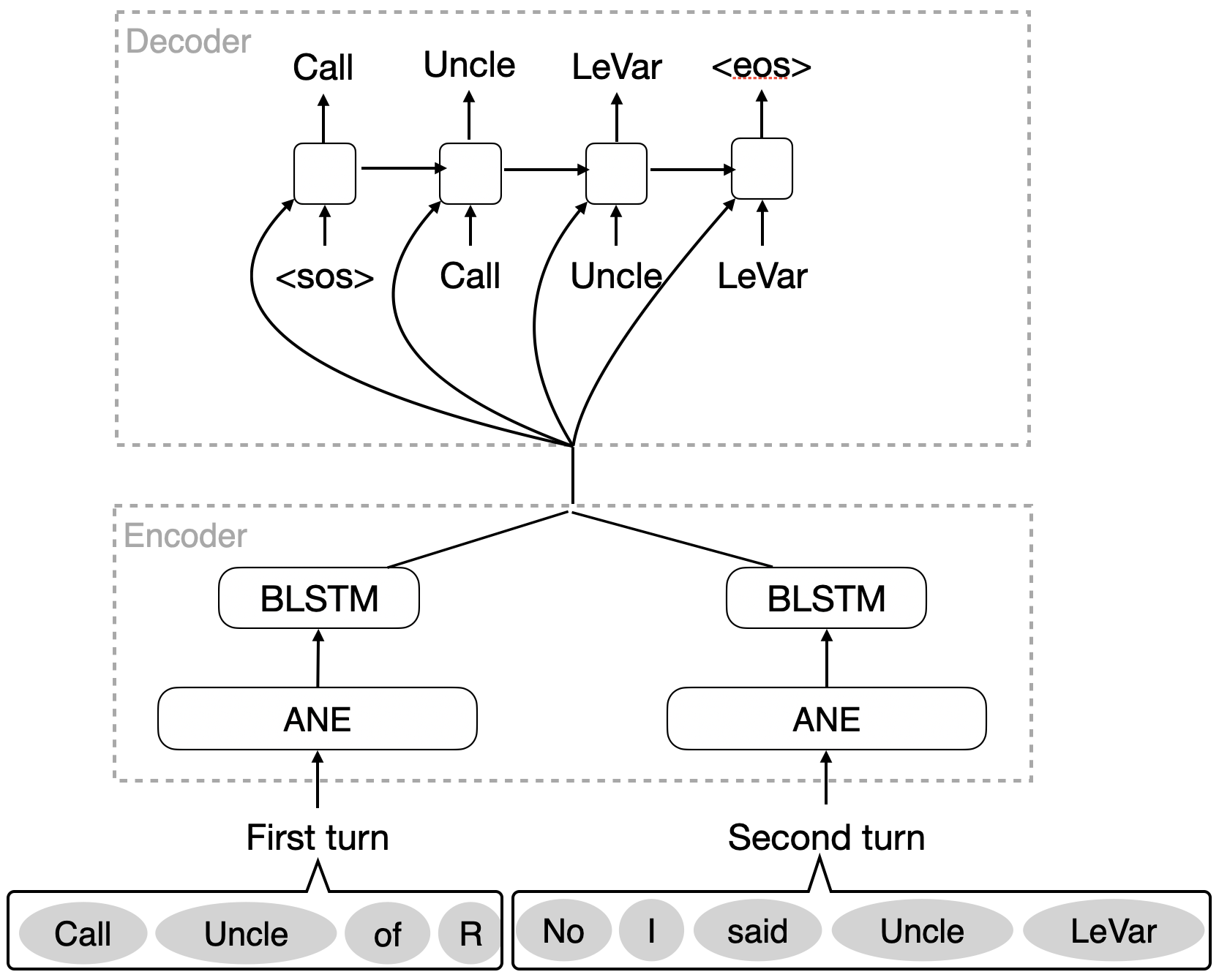}
\caption{A diagram of the proposed model. The words are embedded using the ANE model. These embeddings are then encoded using a BLSTM. The decoder generates the rewritten query one word at a time using a pointer network.}
\label{fig:model}
\end{figure}

Our proposed model takes two utterances as inputs: the potentially incorrect first utterance and the correction followup utterance. The output is a rewritten query. The rewritten query is represented as a sequence of pointers to words in the incorrect first utterance and the second utterance. To clarify take the following example:

\begin{itemize}
\item \textbf{First utterance}: Call Uncle LeVar
\item \textbf{ASR transcription}: Call Uncle of R
\item \textbf{Second utterance}: No, I said Uncle LeVar
\item \textbf{Model output}: 1-1 1-2 2-5
\end{itemize}

\noindent 1-1 represents first word in first utterance (Call), 1-2 represents second word in first utterance (Uncle), 2-5 represents the fifth word in the second utterance, LeVar in this example. The model thus replaces ``of R'' with ``LeVar'', keeping the ``call'' verb. Some words can occur in both sentences, such as ``Uncle''. We select the first-turn word as target during training. Figure \ref{fig:model} depicts a diagram of the proposed model. Each component is discussed in detail in the following subsections.

\subsection{Encoder}
\label{sec:encoder}

We encode each utterance into a sequence of context-aware word vectors. We embed the words into ANEs. The ANE model is an LSTM that takes in the sequence of graphemes that make up the word and produces an embedding vector \cite{jeon2020acoustic}:

\begin{equation}
	\boldsymbol{e} = \text{ANE}(\boldsymbol{G}),
\end{equation}

\noindent where $\boldsymbol{e}$ is the ANE and $ \boldsymbol{G}$ is the grapheme sequence. Finally, we pass the ANEs through a bidirectional LSTM (BLSTM) \cite{hochreiter1997long} to create context-aware word representations.

\subsection{Decoder}

\begin{figure}
\includegraphics[width=0.9\linewidth]{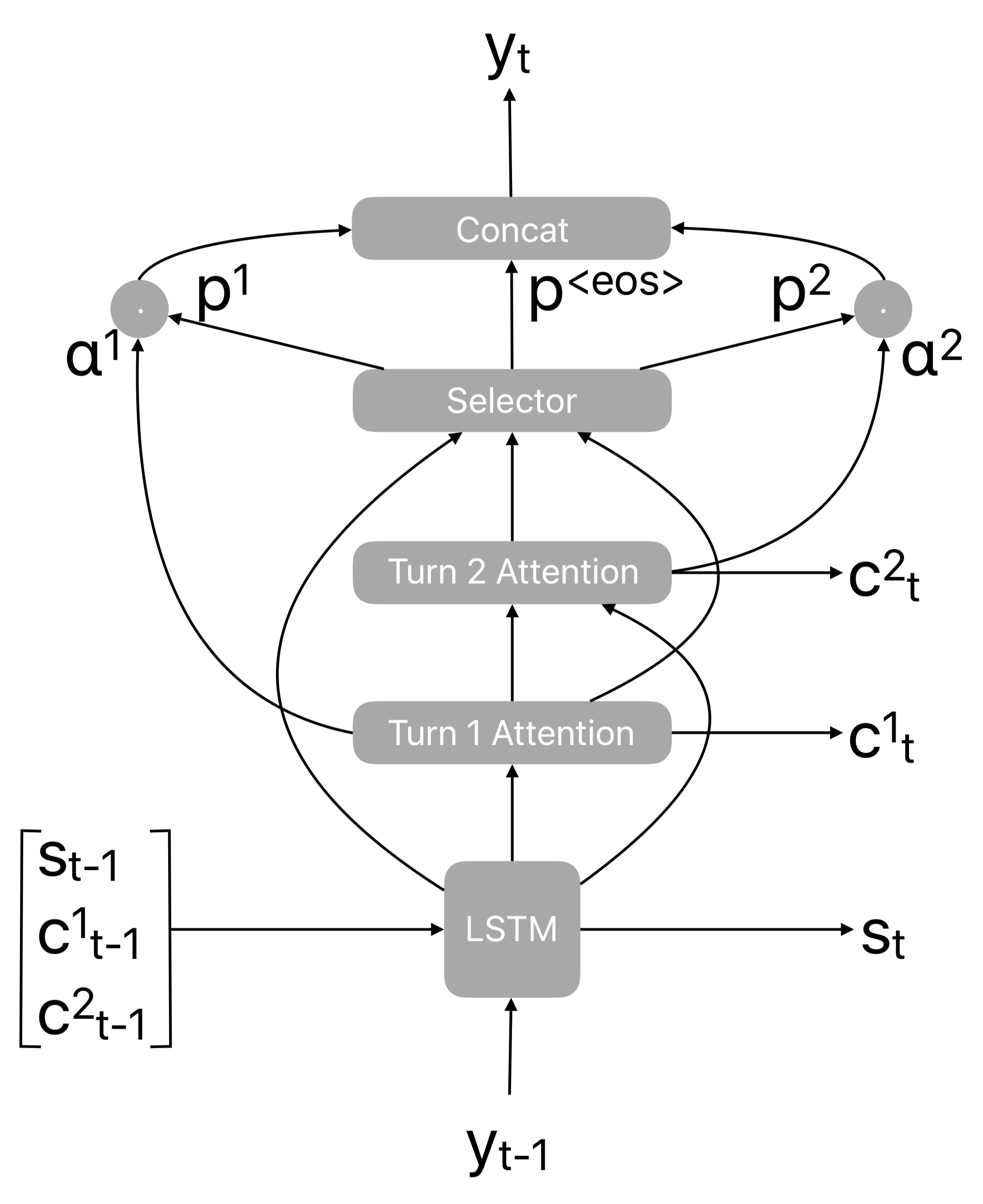}
\caption{A schematic of one step of the 2SA decoder. The LSTM updates the state. The utterance 1 Attention looks in the first utterance for the relevant word. The utterance 2 Attention then looks into the second utterance for a similar sounding alternative. Finally, the selector selects between the two candidates or the end-of-sequence label.}
\label{fig:decoder}
\end{figure}

A schematic of one step of the 2SA decoder is found in Figure \ref{fig:decoder}. The 2SA decoder is a modified pointer network \cite{vinyals2015pointer} and takes the following inputs:

\begin{itemize}
	\item $\boldsymbol{y}_{t-1}$: The word from the previous decoder step. This vector is taken out of the word-level representation discussed in Section \ref{sec:encoder}.
	\item $\boldsymbol{s}_{t-1}$: The decoder LSTM state from the previous decoder step.
	\item $\boldsymbol{c}^1_{t-1}$: The first utterance attention context from the previous decoder step.
	\item $\boldsymbol{c}^2_{t-1}$: The second utterance attention context from the previous decoder step.
\end{itemize}

\noindent The output of the decoder step is the next word of the output sequence, which is chosen from the words in the two input utterances. \\

First, we feed all the inputs into the LSTM to obtain the updated decoder LSTM state $\boldsymbol{s}_{t}$. This updated state is used to query the first utterance using an attention mechanism \cite{bahdanau2014neural}. The attention mechanism gives a vector $\boldsymbol{\alpha}^1$ that contains a weight for each word in the first utterance. These attention weights are used to compute the context vector $\boldsymbol{c}^1_t$, which is a weighted average of the context-aware word vectors using $\boldsymbol{\alpha}^1$ as weights. We then query the second utterance using the context from the first utterance and the LSTM state in the same way to get the attention weights $\boldsymbol{\alpha}^2$ and second utterance context vector $\boldsymbol{c}^2_t$.\\

The intuition of the above is as follows: Both attention mechanisms look into an utterance for a candidate word to output next. We start by looking into the first utterance, because the corrected transcription is expected to be similar to the first utterance, so it is relatively easy to know which part to attend to next. Once we have a  candidate word from the first utterance, we look into the second utterance for a similar sounding word that could act as a replacement.\\

Once the attention mechanisms have found a candidate for each utterance, the selector chooses which word to output. The selector receives the LSTM state and both attention contexts as input and outputs three probabilities:

\begin{enumerate}
	\item $p^1$: Probability that the candidate from the first utterance is correct
	\item $p^2$: Probability that the candidate from the second utterance is correct
	\item $p^{\langle\text{eos}\rangle} $: Probability for ending the sequence
\end{enumerate} 

We then create a probability distribution over all words in both utterances and the $\langle\text{eos}\rangle$-label by multiplying each utterance probability with the respective attention weights:

\begin{equation}
	\boldsymbol{p} = \begin{bmatrix}
		p^1 \boldsymbol{\alpha^1} \\
		p^2 \boldsymbol{\alpha^2} \\
		p^{\langle\text{eos}\rangle} 
	\end{bmatrix},
\end{equation}

\noindent where $\boldsymbol{p}$ is the probability distribution. We then select the next word $\boldsymbol{y}$ from this distribution.\\

\subsection{Inference}

During training we use the ground truth output sequences as inputs to the decoder. For inference we do beam search with beam width 3 to find the most likely output sequence, where we keep generating words until an end-of-sequence token is generated. We also enforce a strict left-to-right attention policy in both attention mechanisms. This means that the attention mechanism can only attend to words to later in the utterance than the last word selected from this utterance.

\subsection{Classifier}

For a virtual assistant, the vast majority of interactions are not repetition-based recoveries. For these cases, the model should not rewrite the original query. To determine whether a rewrite is necessary, we add a classifier that classifies a rewrite as either necessary or not necessary. In this paper, we compute a phonetic weighted Levenshtein distance between the original request and the rewrite. The costs of the edits in the Levensthein distance are based on a phonetic edit distance that was computed based on the confusability of phones in the ASR. If this distance is lower than some threshold, the rewrite is deemed necessary. 

\section{Training data generation}
\label{sec:training}

To train the model, we synthetically generated repetition-based recovery interactions based on a transcribed speech dataset. We first run all the utterances in the dataset through the ASR to get ASR transcriptions. For the cases, where the ASR transcription contains an error we find the substitution the ASR made and generate a repetition-based recovery. In the example in Section \ref{sec:model}, the ASR made the substitution ``LeVar'' $\rightarrow$ ``of R''. Using this substitution, a repetition-based recovery is generated using the following heuristics:

\begin{itemize}
    \item A random number of words to the left and right of the error are included in the recovery. For 85\% of cases, this is 0 words, for 10\% this is one word and for 5\% this is 2 words.
    \item For 10\% of repetition-based recoveries, a prefix is included. This prefix was randomly selected for a list of common prefixes, like ``No, I said''. 
\end{itemize}

With these heuristics some of the repetition-based recoveries that can be generated for the above example are:
\begin{itemize}
    \item LeVar
    \item Uncle LeVar
    \item No, I said uncle LeVaR
\end{itemize}

The model is trained using the ASR transcription and generated repetition-based recovery as input and the reference transcription as the target rewrite. 

\section{Baseline}
\label{sec:baseline}

We use two baselines to compare the results of our proposed model: one rule-based model based on phonetic alignments and one more standard pointer network similar to our proposed model.  

The rule-based model generates a list of all possible n-gram pairs of both utterances for potential replacements. We compute a phonetic edit distance for all pairs and generate the corrected transcription by replacing the n-gram in the first utterance with the n-gram from the correction follow-up. We compute the phonetic edit distance using a phone confusion matrix and normalise that distance by dividing by the number of phones in the first utterance n-gram.

For the example given in Section \ref{sec:model}, the replacement with the smallest phonetic edit distance would be for replacing ``Uncle of R'' with ``Uncle LeVar''.   

Furthermore, we implement an unmodified pointer network \cite{vinyals2015pointer} to compare with our proposed 2SA pointer network. We use the same ANE vectors as input instead of word embeddings.

\section{Experiments}
\label{sec:experiments}
\subsection{Dataset}
Both neural network models are trained on an internally collected dataset described in Section \ref{sec:training}. Using the transcribed first turn, we generate the second turn and the target rewrite. This leads to about 350k turn pairs. For evaluation, we employ a fully annotated test set with 9.2k turn pairs. The annotators transcribe both turns and indicate whether the user intended to correct in the second turn.

\begin{table}[!htbp]
\begin{tabular}{@{}lrr@{}}
\toprule
                   & \multicolumn{1}{l}{count} & \multicolumn{1}{l}{proportion} \\ \midrule
Rewrite correction & 450                       & 4.8\%                          \\
No correction      & 8840                      & 95.2\%                         \\ \bottomrule
\end{tabular}
\caption{In this table we show the evaluation dataset statistics. The human-graded evaluation dataset reflects traffic, where majority of follow-ups are not correction.}
\label{tab:dataset}
\end{table}

\subsection{Metrics}
We compute the word error rate reduction (WERR) by comparing the WER of the original first turn utterances with the WER of the rewrites.

\begin{align}
    WERR = 1 - \frac{WER\_rewrite}{WER\_original}
\end{align}

We measure WERR only on turn pairs that have been annotated as correction, as a rewrite is not required for the others. We additionally compute false alarm rates (FAR) at a range of classifier thresholds. FAR is identical to false positive rate and it's the proportion of unnecessary rewrites. 

\subsection{Training setup}
The baseline pointer network has 1 encoder and decoder BLSTM layer with 128 hidden size. We train it using learning rate 0.0003 with Adam and batch\_size 32. For our proposed model we employ 2 encoders (one for each turn) and 1 decoder with the same dimension, but modified attention head. We train it with batch size 128 and learning rate 0.0001 with Adam. We use pretrained ANEs that are fixed during training.

While the 2SA has twice as many encoder parameters, each encoder only sees half of the data as they only receive their respective turn. For the pointer network, we concatenate both turns to generate a rewrite.

\section{Results}
\label{sec:results}
\begin{figure}[h]
\includegraphics[width=8cm]{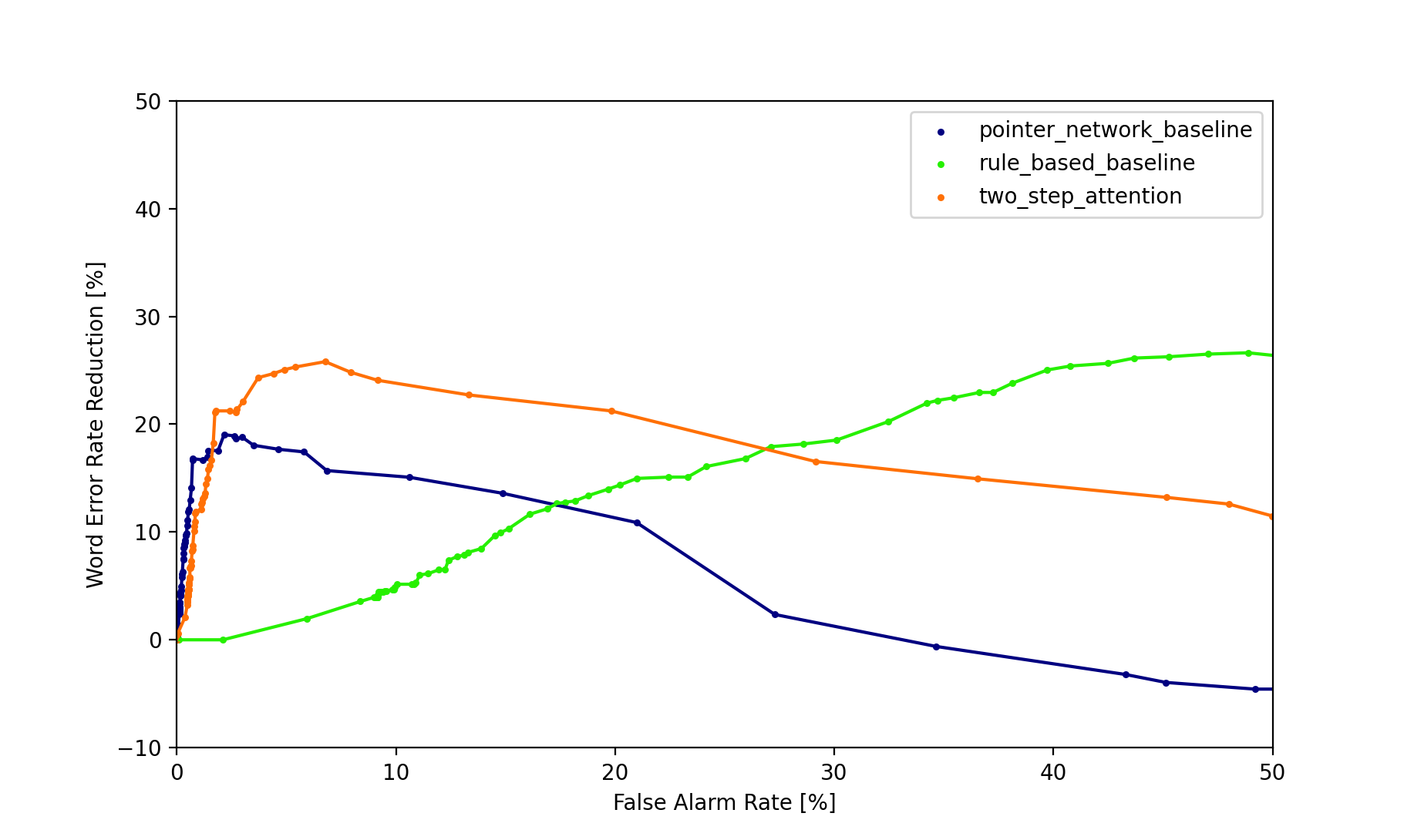}
\caption{On the x-axis we show the False Alarm Rate. On the y-axis we show the Word Error Rate Reduction. (Higher is better) Each point represents a classifier threshold. Low threshold corresponds to high FAR and vice versa. We compare the rule-based baseline with two neural network approaches.}
\label{fig:cbr_plot}

\end{figure}

We measure the WERR of our three proposed approaches on the test set at all classifier thresholds in Figure \ref{fig:cbr_plot}. We zoom in on the WERR results in Table \ref{tab:model_res}.

The rule based baseline achieves a higher maximum WERR, but at a higher FAR. Both machine learned models give high WERR early on before tapering off. We compare the pointer-network and the 2SA network and note the pointer-network has lower FAR for the same WERR, but lower maximum WERR overall, showing a FAR/WERR tradeoff between the two model approaches.

We note the rule-based rewriter uses the same edit distance minimizing rules on positive and negative pairs. Our classifier uses edit distance to score the rewrite. Given the unbalance of our evaluation dataset and real life traffic, it requires a low classifier threshold (implying high FAR) to be feasible. The machine learned do a rewrite when it makes sense to do so, and output random rewrites otherwise. These rewrites will have a high edit distance and be filtered by the classifier.

\begin{table}[]
\begin{tabular}{@{}lrr@{}}
\toprule
Model                    & Max WERR & FAR     \\ \midrule
rule-based baseline      & 26.60\%  & 48.88\% \\
pointer-network          & 19.03\%  & \textbf{2.15}\%  \\
2 step attention pointer-network         & 25.80\%  & 6.77\%  \\ \bottomrule
\end{tabular}
\caption{We evaluate the model performance on the test set. We find the best possible WERR and show its corresponding FAR}
\label{tab:model_res}
\end{table}

\section{Future Work}
\label{sec:future}

The user can try to help correct errors by providing extra information to the system. For example, the user might describe an entity that was misrecognized. This is not captured by pure query rewrite. In future work, one could investigate modifying the task to incorporate this extra information.\\

The proposed system cannot correct the error if the ASR repeats the error in the correction followup. However, it is possible that the correct hypothesis is present in the n-best list of the speech recognizer. By combining information from both utterances, it might be possible to surface the correct hypothesis.\\

\section{Conclusion}
\label{sec:conclusion}

We propose a system that allows the user to correct recognition errors through repetition. The model handles this repetition-based recovery by rewriting the original query to incorporate the follow-up correction. The rewritten query is then scored with an edit distance based classifier for thresholding. We propose a 2-Step Attention pointer network and show that it outperforms both a standard pointer network and a rule-based baseline. We propose a way to generate synthetic training data using only transcribed data and an ASR system to train our machine learned models. 

\bibliographystyle{IEEEtran}

\bibliography{mybib}

\end{document}